\let\IEEEAccessSavedYear\year
\documentclass{ieeeaccess}
\let\year\IEEEAccessSavedYear

\usepackage{cite}
\usepackage{amsmath,amssymb,amsfonts}
\usepackage{silence}
\usepackage{amsthm} 
\usepackage{mathtools}
\usepackage{siunitx}
\usepackage{cuted}
\usepackage{graphicx}
\usepackage{subcaption}
\usepackage{epstopdf}
\usepackage[hidelinks]{hyperref}
\usepackage{tabularx}
\usepackage{multirow}
\usepackage{booktabs}
\usepackage{diagbox}
\usepackage{cuted}
\usepackage{lipsum}
\newcommand{\includestandalonedefaultmode}{buildnew}%
\usepackage[group=true,mode=\includestandalonedefaultmode]{standalone}%

\usepackage[shortcuts,acronym,automake]{glossaries}
\makeglossaries
\newacronym{6g}{6G}{Sixth Generation}
\newacronym{ai}{AI}{artificial ntelligence}
\newacronym{dt}{DT}{digital twin}
\newacronym{si}{SI}{swarm intelligence}
\newacronym{iot}{IoT}{Internet of Things}
\newacronym{iiot}{IIoT}{Industrial Internet of Things}
\newacronym{pso}{PSO}{Particle Swarm Optimization}
\newacronym{uav}{UAV}{unmanned aerial vehicle}
\newacronym{mec}{MEC}{Multi-Access Edge Computing}
\newacronym{vfh}{VFH}{Vector Field Histogram}
\newacronym{per}{PER}{packet error rate}
\newacronym{p2p}{P2P}{peer-to-peer}
\newacronym{ul}{UL}{uplink}
\newacronym{dl}{DL}{downlink}
\newacronym{agv}{AGV}{Automated Guided Vehicle}
\newacronym{gbest}{G-Best}{group best position}
\newacronym{X}{X-ULK}{unknown leak source}

\usepackage[english]{babel}
\usepackage[utf8]{inputenc}
\usepackage{algorithm}
\usepackage{algpseudocode}
\usepackage{float}
\usepackage{amsmath}

\usepackage{graphicx}
\usepackage{threeparttable} 

\usepackage[normalem]{ulem}
\newcommand{\revise}[1]{{\color{blue}#1}}
\renewcommand{\revise}[1]{{#1}}

\newif\ifreviewmode
\reviewmodetrue 

\ifreviewmode
\else
\renewcommand{\todo}[1]{} 
\renewcommand{\revise}[2]{#2} 
\fi

\usepackage{textcomp}

\usepackage{bm}
\makeatletter
\AtBeginDocument{\DeclareMathVersion{bold}
	\SetSymbolFont{operators}{bold}{T1}{times}{b}{n}
	\SetSymbolFont{NewLetters}{bold}{T1}{times}{b}{it}
	\SetMathAlphabet{\mathrm}{bold}{T1}{times}{b}{n}
	\SetMathAlphabet{\mathit}{bold}{T1}{times}{b}{it}
	\SetMathAlphabet{\mathbf}{bold}{T1}{times}{b}{n}
	\SetMathAlphabet{\mathtt}{bold}{OT1}{pcr}{b}{n}
	\SetSymbolFont{symbols}{bold}{OMS}{cmsy}{b}{n}
	\renewcommand\boldmath{\@nomath\boldmath\mathversion{bold}}}
\makeatother

\def\BibTeX{{\rm B\kern-.05em{\sc i\kern-.025em b}\kern-.08em
		T\kern-.1667em\lower.7ex\hbox{E}\kern-.125emX}}

\begin{document}
	\history{Date of publication xxxx 00, 0000, date of current version xxxx 00, 0000.}
	\doi{10.1109/ACCESS.2025.xxxxxx}
	
	\title{A Communication-Efficient Digital Twin Framework for PSO-Based Swarm Navigation and Obstacle Avoidance}
	\author{\uppercase{Siyu~Yuan}\authorrefmark{1}, \uppercase{Khurshid~Alam}\authorrefmark{2}, \uppercase{Dennis~Krummacker}\authorrefmark{2},\\
		\uppercase{Bin~Han}\authorrefmark{1}~\IEEEmembership{Senior Member, IEEE},
		and \uppercase{Hans~D.~Schotten}\authorrefmark{1,2}~\IEEEmembership{Member, IEEE}
	}
	
	\address[1]{RPTU University Kaiserslautern-Landau, 67663 Kaiserslautern, Germany}
	\address[2]{German Research Center for Artificial Intelligence (DFKI), 67663 Kaiserslautern, Germany}
	
	\tfootnote{This work was supported in part by the Federal Ministry of Research, Technology and Space (BMFTR) of Germany in the project Open6GHub+ under Grant 16KIS2406.}
	
	\markboth
	{Yuan \headeretal: A Communication-Efficient Digital Twin Framework for PSO-Based Swarm Navigation and Obstacle Avoidance}
	{Yuan \headeretal: A Communication-Efficient Digital Twin Framework for PSO-Based Swarm Navigation and Obstacle Avoidance}
	
	\corresp{Corresponding author: Bin Han (e-mail: bin.han@rptu.de).}

	\begin{abstract}
		Swarm-based target localization in industrial environments faces two major challenges: navigating obstacle-rich spaces and managing intensive communication among agents. This paper proposes a communication-efficient \ac{dt} framework for \ac{pso}-based swarm navigation and obstacle avoidance. The \ac{dt}, deployed on a \ac{mec} server, maintains a virtual replica of the environment to provide global guidance and obstacle bypassing when agents become trapped or experience poor connectivity. By reducing unnecessary peer-to-peer communication and centralizing environmental information, the proposed framework improves both navigation efficiency and communication resource utilization. Simulation results demonstrate that the \ac{dt}-assisted \ac{pso} with obstacle avoidance achieves faster convergence and significantly lower communication load compared with decentralized P2P and random-walk \ac{pso} approaches. These findings highlight the potential of integrating \ac{dt} with swarm intelligence to enhance cooperative exploration in complex industrial scenarios such as chemical leakage localization.
	\end{abstract}
	
	
	\titlepgskip=-21pt
	
	\maketitle
	
	\glsresetall
	
	\section{Introduction}\label{sec:intro}
	\PARstart{I}{n} contemporary industrial scenarios, the imperative for robust multi-device connectivity and seamless communication is increasingly critical. The next phase of industrial advancement necessitates more powerful and modern communication methodologies, with \ac{6g} ~\cite{9349624,chowdhury20206g} poised to play a pivotal role in this evolution. In particular, \ac{6g} helps to provide ubiquitous coverage and support massive access for a wide range of devices and applications, meeting the growing demands of modern industrial environments. The continuous evolution of network communication technology indicates that \ac{6g}, with its unprecedented capacity, ultra-low latency, native intelligence, and enhanced security, has the potential to integrate with various emerging technologies, such as \ac{dt} technology~\cite{khan2022digital,han2023digital}, machine learning, \ac{mec}, and Distributed Ledger Technology, etc. These integrations, facilitated by \ac{6g}, will create substantial advancements in various industries. These capabilities make \ac{6g} a pivotal enabler of industrial-scale applications, particularly in scenarios that demand extensive coverage, massive device connectivity, and real-time synchronization.
	
	One of the most promising applications of this integration is \ac{dt} ~\cite{9923927, DenKr_DigitalTwin_MathModel}, which leverages real-time data processing, communication, and synchronization to optimize operations and decision-making in complex environments. With \ac{dt} technology, various features, states, and behaviors of physical objects can be accurately simulated and reflected in digital space. 
	On the other hand, as a branch of artificial intelligence, \ac{si} studies the collective intelligence emerging from groups of simple agents~\cite{bonabeau1999swarm,martens2011editorial}.
	It exhibits a notable ability to share local information and experiences among swarm members, which improves the performance along with the swarm size. When empowered by \ac{6g} networks, \ac{si} can fully exploit its scalability, as \ac{6g}'s massive connectivity and high reliability ensure efficient communication even in densely deployed scenarios. However, in wireless scenarios, this advantage may be compromised by the limited channel capacity, especially when the deployment is dense. The work in~\cite{YHKS2022massive} compares the communication efficiency of the \ac{dt} approach against device-to-device communication methods. It has been demonstrated that combining \ac{dt} and \ac{si} can enhance communication efficiency by reducing overhead and improving data relevance. By synchronizing the agents' information to their \acp{dt} aggregated on a \ac{mec} server, and virtualizing the information exchange among agents, the bottleneck of air interface can be well resolved for \ac{si}. This virtualized information exchange mechanism is particularly suited to \ac{6g}-enabled architectures, where \ac{mec} and \ac{dt} technologies are natively supported and optimized.
	
	Investigating an \ac{uav} localization problem, the study of~\cite{YHKS2022massive} oversimplified the real world scenario by neglecting the presence of obstacles, a common factor in practical industrial settings, if not inevitable. Similar limitations appear in other swarm navigation studies. For instance, the work in~\cite{mcguire2019minimal} introduces a minimal navigation solution for a swarm of miniature flying robots to cooperatively explore an unknown environment, yet it does not fully exploit the potential of communication reuse among agents, leading to suboptimal swarm efficiency. The integration of \ac{dt} technology can compensate for this limitation by enabling synchronized virtual interaction and efficient information reuse. Likewise, the study of~\cite{ducatelle2014cooperative} proposes a cooperative navigation strategy where each agent relies on random encounters with others to obtain local information during navigation, but its performance depends heavily on continuous interaction opportunities, which may not always exist in structured or obstacle rich environments, thereby constraining target localization efficiency. Moreover, the study of~\cite{YHKS2022massive} adopts the classical \ac{pso} ~\cite{kennedy1995particle}, which is inadequate for direct application in obstacle laden scenarios. While several traditional obstacle avoidance algorithms, such as the A* algorithm~\cite{DP1985generalized} and Dijkstra’s algorithm~\cite{Dijkstra1959note}, have been well developed and widely applied, they are predominantly designed for independent decision making of individual robots. Hence, developing novel \ac{si} oriented obstacle avoidance approaches and leveraging the advantages offered by \ac{dt} emerges as an intriguing and necessary research direction.
	
	
	To address the aforementioned limitations, this paper proposes a communication efficient framework that integrates \ac{pso} with \ac{dt} technology. Each swarm agent is digitized into a virtual twin that models both its behavior and an obstacle rich environment in real time. This architecture allows agents to exchange essential state information through the \ac{dt} layer rather than frequent peer-to-peer communication, significantly reducing network overhead. Moreover, the \ac{dt} assisted obstacle avoidance mechanism enables agents to maintain efficient navigation and avoid deadlocks when physical obstacles or poor connectivity occur. As a result, the swarm achieves faster convergence toward unknown target points with lower communication cost. By synchronizing the virtual and physical environments in real time, the proposed method improves swarm coordination, enhances navigation efficiency, and accelerates target localization in complex industrial environments.
	
	The remainder of this paper is organized as follows. Sec.~\ref{sec:setup} presents the problem setup. Sec.~\ref{sec:approaches} describes the proposed system framework. Sec.~\ref{sec:simulations} presents the simulation setup and numerical results. Sec.~\ref{sec:discussions} provides further discussion, and Sec.~\ref{sec:conclusion} concludes the paper.

	\section{Problem Setup}\label{sec:setup}
	\begin{figure}[!htpb]
		\centering
		\includegraphics[width=0.9\linewidth]{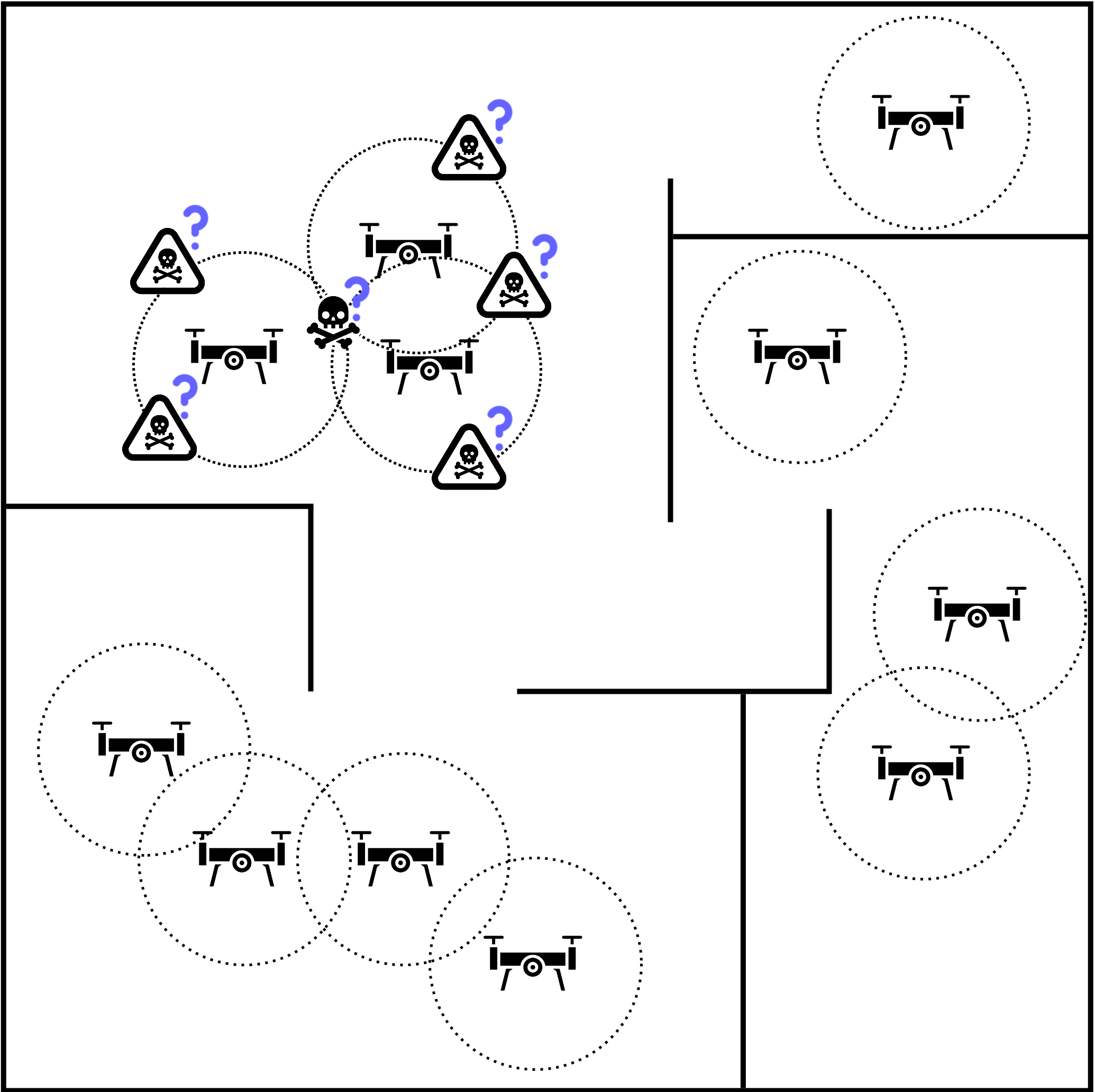}
		\caption{Scenario of the studied problem}
		\label{fig:ProbScene}
	\end{figure}
	The application scenarios in this paper focus on emergency response for chemical leakages in production facilities. In such events, hazardous substances spread rapidly and unpredictably, creating immediate safety risks. This scenario presents two critical challenges: first, accurately locating the leak source, and second, achieving this identification within strict time constraints. While each \ac{uav} is equipped with concentration sensors to detect proximity to the source, precisely pinpointing the leak origin under time pressure remains a significant unresolved problem. The structural layout of chemical plants, containing numerous fixed walls and machinery, creates an obstacle rich environment that severely limits UAV accessibility. As illustrated in Fig. ~\ref{fig:ProbScene}, these obstacles prevent many \ac{uav}s from participating in critical detection tasks, with only a subset of the swarm able to reach and accurately identify the target site. The speed of leak source identification is directly related to the \ac{uav}s' ability to mitigate hazards and minimize economic consequences, making rapid navigation through complex indoor environments with limited visibility and accessibility a paramount concern.
	
	\ac{si} algorithms like PSO are well suited for the localization problem due to their distributed problem solving capabilities. Since only when multiple agents collaborate can they achieve more precise localization, marking, and containment of the leak point, PSO is particularly effective in facilitating this convergence.  However, the structural layout of chemical plants, often containing numerous fixed walls and other machinery, complicates the task of locating the leak source. This can cause \ac{uav}s to become trapped near barriers, significantly delaying convergence and compromising the time critical nature of emergency response.
	
	Rapid convergence to the target location often requires obstacle avoidance algorithms. However, to effectively balance the dual objectives of avoiding obstacles and approaching the target, virtually all mainstream obstacle avoidance algorithms require prior knowledge of the target position. This presents a fundamental contradiction, as the problem addressed in this paper specifically involves scenarios where the target location is unknown.
	
	In this paper, we propose a potential solution to address this challenge.

	\section{System Framework}\label{sec:approaches}

	\begin{figure*}[!htpb]
		\centering
		\includegraphics[width=\linewidth]{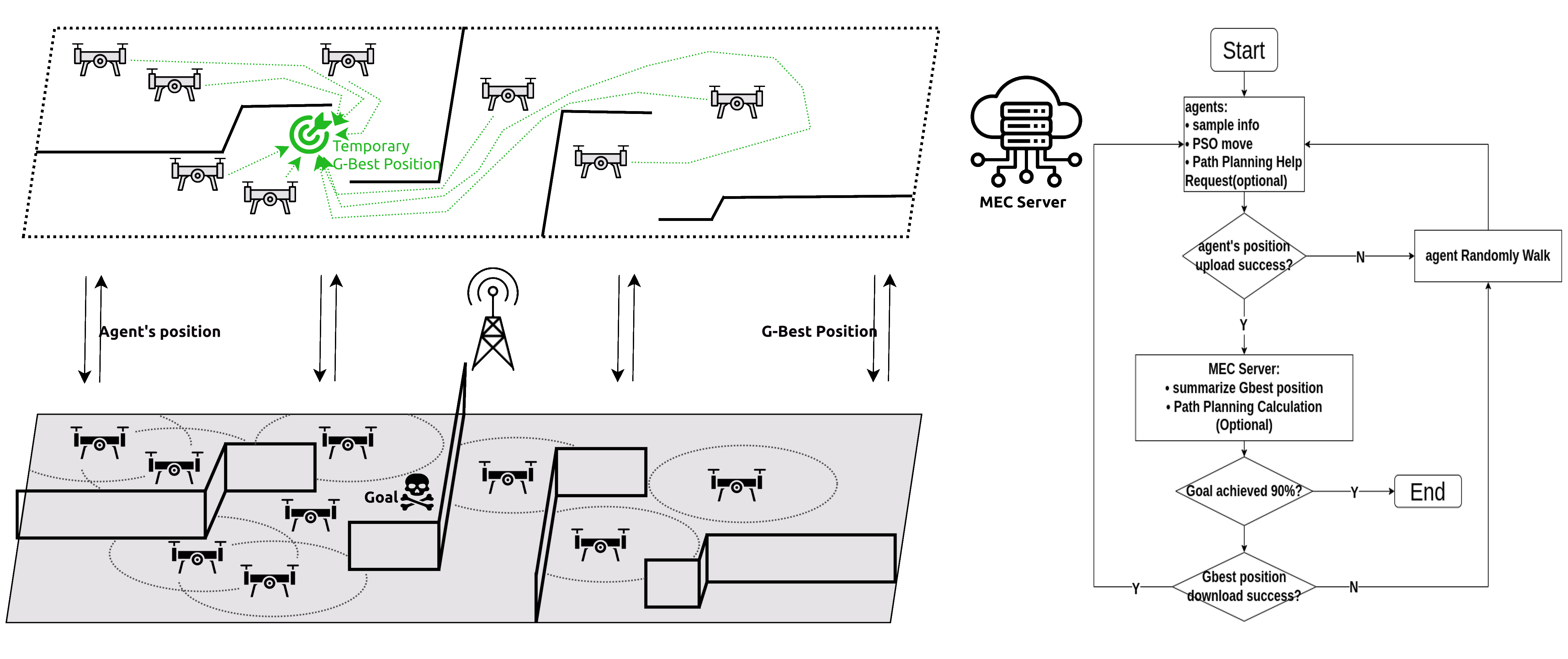}
		\caption{The proposed \ac{dt}-assisted \ac{pso} framework. The left part illustrates the physical environment and UAV agents operating collaboratively under DT-based synchronization, while the right part presents the workflow of the proposed coordination process.}
		
		\label{fig:Scenedt}
	\end{figure*}
	\revise{
		To address the challenges of obstacle constrained navigation and unknown leak source, this work proposes a \ac{dt} assisted swarm coordination framework integrating \ac{pso}, MEC-based synchronization, and obstacle aware rescue mechanisms. As illustrated in Fig.~\ref{fig:Scenedt}, the proposed framework consists of three major components: the \ac{mec} server, a swarm of \acp{uav}, and the base station. The MEC server maintains digital twins of both UAV agents and environmental obstacles, enabling centralized coordination and obstacle aware assistance. Each UAV estimates its distance to the unknown leak source using onboard sensing modules and exchanges coordination information with the MEC server through the base station. 
	}
	\revise{
		\subsection{System Model}
	}
	
	\revise{
		The operational environment is modeled as a two dimensional grid map denoted by $\mathcal{E}$. The environment consists of obstacle free regions and obstacle regions, expressed as
		
		\begin{equation}
			\mathcal{E}=\mathcal{F}\cup\mathcal{O},
		\end{equation}
		
		where $\mathcal{F}$ denotes the free space and $\mathcal{O}$ represents the obstacle space. UAV agents are only allowed to move within feasible regions belonging to $\mathcal{F}$.
	}
	
	\revise{
		At iteration $t$, the state of UAV $i$ is defined as
		
		\begin{equation}
			s_i^t=
			\left(
			p_i^t,
			v_i^t,
			d_i^t,
			c_i^t
			\right),
		\end{equation}
		
		where $p_i^t\in\mathbb{Z}^2$ denotes the UAV position on the grid map, $v_i^t$ represents the velocity vector, $d_i^t$ denotes the estimated Euclidean distance between UAV $i$ and the unknown leak source, and $c_i^t\in\{0,1\}$ indicates the communication state, where $c_i^t=1$ denotes successful communication and $c_i^t=0$ indicates communication failure. The grid-based representation is adopted to simplify obstacle occupancy validation and feasible path generation.
	}
	
	\revise{
		Each UAV is equipped with onboard concentration sensing modules capable of estimating the Euclidean distance to the unknown leak source according to
		\begin{equation}
			d_i^t=
			\|p_i^t-X\|,
		\end{equation}
		where $X$ denotes the unknown leak source location. During each iteration, the MEC hosted DT infrastructure synchronizes the received UAV states and obstacle information to maintain a global representation of the operational environment. To account for synchronization latency between physical UAVs and their digital twins, the DT-maintained position of UAV $i$ is modeled as
		\begin{equation}
			\hat{p}_i^t = p_i^{t-\tau_i},
		\end{equation}
		where $\hat{p}_i^t$ denotes the position of UAV $i$ maintained by the MEC hosted DT at iteration $t$, and $\tau_i$ represents the synchronization delay of UAV $i$. The synchronization delay is assumed to be bounded within a finite coordination interval.
	}
	
	\revise{
		\subsection{DT Assisted PSO Coordination}
	}
	
	\revise{
		As illustrated in the workflow shown in Fig.~\ref{fig:Scenedt}, the proposed DT assisted coordination process operates iteratively. During each iteration, UAV agents estimate their distances to the unknown leak source using onboard sensing modules and attempt to upload their current states $(p_i^t,d_i^t)$ to the MEC server through the base station.
		
		Based on the successfully received and synchronized DT states, the MEC server determines the current global best position according to
		\begin{equation}
			gbest^t = \hat{p}_k^t,
			\quad
			k=\arg\min_i \hat{d}_i^t,
		\end{equation}
		where $\hat{d}_i^t$ denotes the distance measurement of UAV $i$ synchronized to the MEC hosted DT at iteration $t$.
		
		The updated global best position is subsequently distributed to UAV agents with successful downlink communication.
	} 
	
	\revise{
		After receiving the updated global best information, each UAV applies the standard \ac{pso} update rule to determine its next movement. In \ac{pso}, the velocity update is jointly guided by three components: the inertia term, the cognitive term associated with the UAV's personal best position, and the social term associated with the global best position. In this work, an additional stochastic perturbation term is included to maintain exploration diversity. The velocity is updated as
		\begin{equation}
			v_i^{t+1}
			=
			wv_i^t
			+
			c_1r_1(pbest_i-p_i^t)
			+
			c_2r_2(gbest^t-p_i^t)
			+
			\xi_i^t,
		\end{equation}
		where $w$ denotes the inertia weight, $c_1$ and $c_2$ represent the cognitive and social coefficients, respectively, $r_1$ and $r_2$ are uniformly distributed random variables within $[-1,1]$, $pbest_i$ is the historical best position of UAV $i$, and $\xi_i^t$ denotes a stochastic perturbation term introduced to improve swarm diversity and reduce premature convergence.
		
	} \revise{
		The UAV position is subsequently updated according to
		\begin{equation}
			p_i^{t+1}
			=
			p_i^t+v_i^{t+1}.
		\end{equation}
		To satisfy practical UAV mobility limitations, the movement velocity is constrained by
		\begin{equation}
			\|v_i^t\|\leq v_{\max},
		\end{equation}
		where $v_{\max}$ denotes the maximum UAV movement capability during one iteration.
	} 
	
	\begin{algorithm}[!t]
		\small
		\caption{DT Assisted PSO-Based Swarm Navigation with Obstacle Avoidance}
		\label{alg:dt_pso_navigation}
		\begin{algorithmic}[1]
			\State \textbf{Input:} Initial positions $\{p_i\}$, obstacle DT map, synchronization delays $\{\tau_i\}$, and hidden source $X$ used only for sensor simulation
			\State \textbf{Initialize:} MEC server, agent digital twins, and PSO parameters
			\While{termination condition is not satisfied}
			\For{each agent $i$ in swarm}
			\State Observe $d_i^t$ using the onboard sensor ($X$ not observable to the agent) and attempt to \textbf{upload} $(p_i^t, d_i^t)$ to MEC server
			\If{upload fails}
			\State Continue local exploration using previously available swarm information; enter random-walk mode if local movement becomes infeasible
			\EndIf
			\EndFor
			\If{MEC receives data from any agents}
			\State Synchronize delayed DT states $(\hat{p}_i^t,\hat{d}_i^t)$ according to $\tau_i$, identify agent $k$ with $\min(\hat{d}_i^t)$ as the current global best agent, and distribute the updated global best position $\hat{p}_k^t$ to all agents with successful downlink
			\EndIf
			\For{each agent $i$}
			\If{downlink received successfully}
			\State Compute a candidate position via the PSO rule using the received global-best information
			\If{the candidate waypoint lies inside the obstacle region}
			\State Request \textbf{assistance} from MEC; MEC generates and the agent follows an obstacle-DT detour to resume exploration
			\EndIf
			\Else
			\State Update a candidate position via local PSO update using the locally cached global-best information; enter random-walk mode if the resulting movement is infeasible
			\EndIf
			\State Validate the candidate position
			\If{the candidate position is invalid}
			\State Attempt to repair the candidate position; restore the last valid position if the repair fails
			\EndIf
			\EndFor
			\EndWhile
			\State \textbf{Output:} Final positions of agents and identified leak source location
		\end{algorithmic}
	\end{algorithm}
	
	\revise{
		\subsection{Communication Failure and Random Walk}
	}\revise{
		Due to signal attenuation and environmental obstacles, both uplink and downlink communication failures may occur during swarm operations.
		
		When uplink communication fails, the MEC server cannot receive the latest UAV state information. Consequently, the corresponding UAV continues local exploration using previously available swarm information. Similarly, when downlink communication fails, the UAV continues executing local PSO updates using the most recently available global best information.
		
		To prevent UAV stagnation in communication degraded regions, the proposed framework employs an obstacle constrained random walk strategy. Instead of remaining stationary, UAVs continue exploring feasible neighboring positions sampled within a bounded movement region constrained by the maximum UAV velocity.
		
		The sampled random walk position is restricted to obstacle free regions of the environment, ensuring that UAV agents avoid invalid movements into obstacle areas.
		
		This mechanism maintains continuous exploration capability even under temporary communication failures, thereby improving the probability of reconnecting with the MEC server and escaping poor communication regions.
		
		In addition, when a UAV reaches the current global best region without successfully locating the leak source, the UAV is also allowed to continue exploration through the random walk mechanism, thereby maintaining swarm diversity and reducing premature convergence.
	}
	
	\revise{
		\subsection{Trap Detection and DT Assistance}
		Environmental obstacles may cause locally generated PSO waypoints to fall inside obstacle regions, resulting in UAV trapping situations. In the proposed framework, a trapped condition occurs when the next waypoint generated by the local PSO update lies inside an obstacle region. Formally, the trap indicator of UAV $i$ is defined as
		\begin{equation}
			\mathrm{Trap}_i^t =
			\begin{cases}
				1, & p_i^{t+1}\in\mathcal{O},\\
				0, & p_i^{t+1}\in\mathcal{F}.
			\end{cases}
		\end{equation}
		
		Upon detecting a trapped condition, the UAV transmits an assistance request to the MEC server whenever communication is available. If communication remains unavailable, the UAV continues local exploration through the random walk mechanism until communication is restored. The MEC hosted DT module then performs a lightweight sampling-based detour planning strategy inspired by probabilistic roadmap (PRM) methods.
		
		Instead of constructing a dense global roadmap, the proposed method incrementally connects feasible sampled waypoints according to obstacle free visibility constraints, thereby reducing computational overhead while enabling rapid rescue path generation.
		
		The generated detour trajectory guides the trapped UAV toward the current global best region while avoiding obstacle collisions. Through iterative execution of this process, the swarm gradually converges toward the unknown leak source. This DT assisted rescue mechanism improves swarm adaptability in obstacle dense environments while preserving collaborative exploration efficiency.
	}
	
	\subsection{\revise{Communication Complexity Analysis}}
	
	\revise{
		Assume that the swarm contains $n$ UAV agents. In the proposed DT assisted framework, each UAV performs one uplink transmission and one downlink reception during each iteration. Therefore, the communication complexity can be expressed as
		\begin{equation}
			C_{DT}=2n,
		\end{equation}
		which yields
		\begin{equation}
			C_{DT}=O(n).
		\end{equation}
		The analysis considers the baseline DT synchronization process during one coordination iteration.
		
		In contrast, conventional peer-to-peer swarm coordination requires each UAV to exchange information with all other UAVs. The corresponding communication complexity becomes	
		\begin{equation}
			C_{P2P}=n(n-1),
		\end{equation}	
		resulting in	
		\begin{equation}
			C_{P2P}=O(n^2).
		\end{equation}
		Therefore, the proposed DT assisted coordination framework significantly reduces communication overhead and improves scalability for large scale swarm operations.
		
		By virtualizing swarm coordination through DT assisted synchronization, the proposed framework enables scalable swarm collaboration while reducing direct inter agent communication dependency.
	}

	\section{Simulation}\label{sec:simulations}

	\subsection{Simulation Setup}
	\begin{figure*}[!htpb]
		\centering
		\includegraphics[width=.9\linewidth]{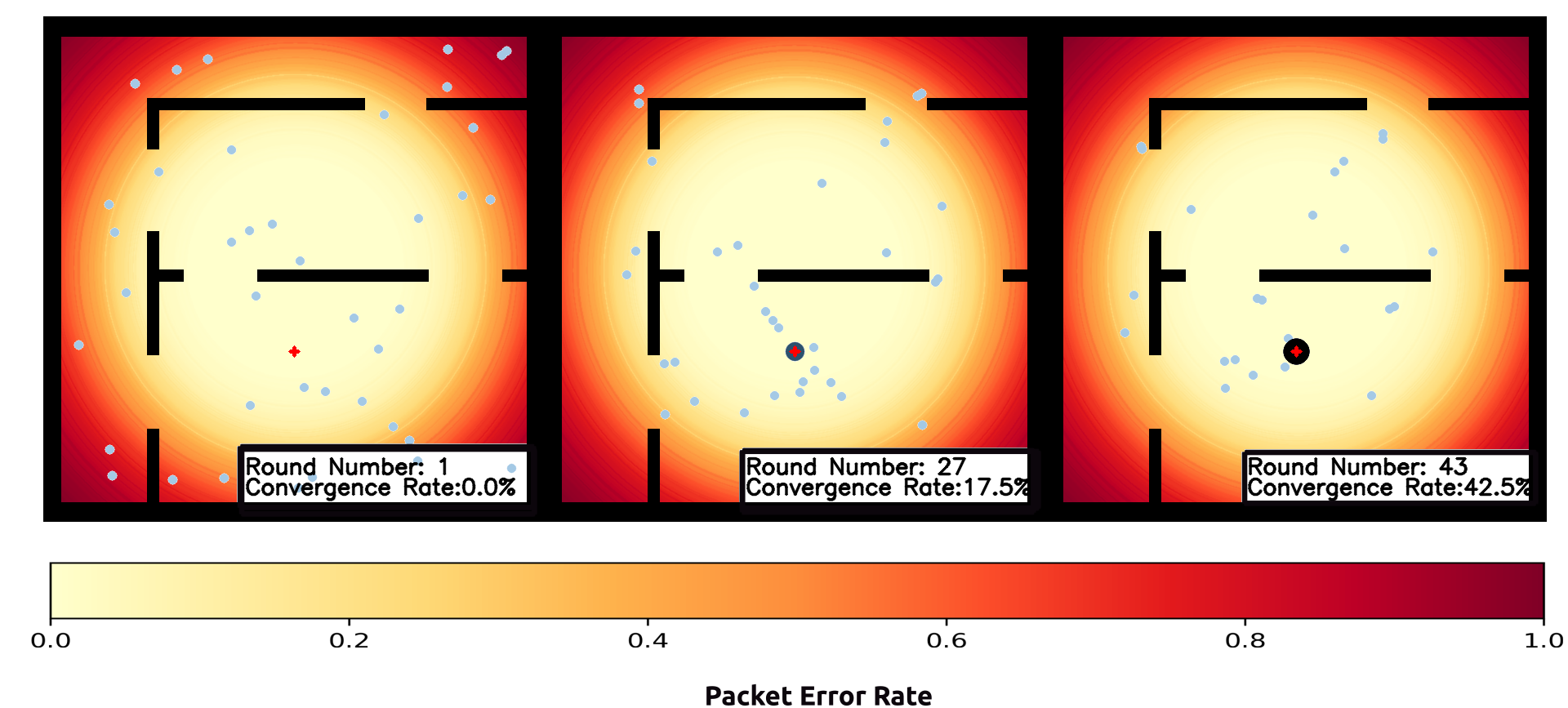}
		\caption{Convergence process of the swarm in an obstacle rich environment: here we illustrate three representative phases of the swarm convergence process, from initial random distribution to near optimal formation. 
			The background color represents the \ac{per}, with red indicating higher error probability and yellow indicating lower.}
		\label{fig:SimGround}
	\end{figure*}
	
	The process illustrated in Fig. \ref{fig:SimGround} depicts the gradual convergence of multiple agents. They move from random initial positions towards an unknown leak source. This process unfolds within a map occupied by obstacles, progressing from left to right across the three maps.
	
	\revise{
		The map is sized at $600\times600$ m and acts as the operational area where agents move toward the leak source. The environment is represented as a grid map containing both obstacle free and obstacle occupied regions. The obstacle occupancy ratio is computed as the number of obstacle cells divided by the total number of grid cells. In the simulated map, the obstacle occupancy ratio is approximately 13.71\%.
	}The red cross, slightly offset from the map's center, denotes the leak source.
	
	The \ac{per} for the map is calculated in ns-3 by positioning the base station at the center and employing the 3GPP propagation loss model. In the \ac{per} Map, lighter shades denote regions with better signal quality, while darker shades indicate relatively poorer signal quality.

	The initial positions of the agents are randomly distributed. Agents in the map follow specific rules, such as a maximum speed limit of 5 meters per second, and how to respond to data transmission failures. Each agent is equipped with two sensors, one of which is used to estimate the distance to the unknown leak source, and the other is used to detect the surrounding environment to avoid obstacles. \revise{Additionally, the sensor used for distance measurement is perturbed with bounded uniformly distributed random noise to simulate imperfect sensing conditions.
	}
	
	\subsection{Numerical Results}
	
	The simulation results encompass five distinct operational modes: P2P mode, DT mode 1, DT mode 2, random walk mode 1, and random walk mode 2. P2P mode is used as the baseline for comparison.
	
	The difference lies in the communication methods: in P2P mode, agents rely entirely on physical world interactions without digital twin infrastructure, requiring each agent to interact with every other agent in the swarm to obtain the \ac{gbest}, resulting in pairwise exchanges among all participants. For instance, in P2P mode, if a swarm contains $n$ agents and each agent aims to obtain the swarm's \ac{gbest}, it would require $n(n-1)$ transmissions in the uplink/downlink. In contrast, in both DT and random walk mode 1, only $n$ transmissions in the uplink/downlink are needed. 
	
	The distinction between \ac{dt} mode 1 and mode 2 lies in their respective transmission strategies and connectivity handling mechanisms. In mode 1 each agent performs only one upload and one download operation per round. \ac{dt} mode 2 is specifically designed to provide a more equitable comparison with P2P mode by incorporating similar transmission attempt frequencies. In \ac{dt} mode 1, when an agent loses network connectivity, it first continues local PSO exploration using the most recently cached global-best information. If the resulting movement is infeasible, it falls back to a random-walk step and waits for the next round's upload and download opportunity.  Conversely, in mode 2, agents remain stationary upon connection failure and perform multiple retransmission attempts. For a swarm of $n$ agents, each failed transmission allows up to $n-2$ additional retry attempts until either successful transmission is achieved or all $n-2$ retry attempts are exhausted. This retry mechanism mirrors the $n-1$ communication attempts required per agent in P2P mode, enabling fair performance comparison between the two approaches.
	
	Random walk modes 1 and 2 use analogous operational differences to their DT counterparts regarding transmission strategies. However, random walk and DT modes differ fundamentally in their operational mechanisms. DT mode integrates PSO optimization with path planning capabilities, leveraging the comprehensive digital twin infrastructure within MEC servers that encompasses both agent and obstacle digital twin modules. In contrast, random walk mode assumes agents are equipped solely with physical world obstacle avoidance sensors. When an agent becomes trapped by obstacles due to PSO algorithm limitations, it can only employ random walk strategies to attempt obstacle circumvention, lacking access to the integrated digital twin framework for coordinated navigation.
	
\begin{figure*}[!htpb]
		\centering
		\includegraphics[width=.9\linewidth]{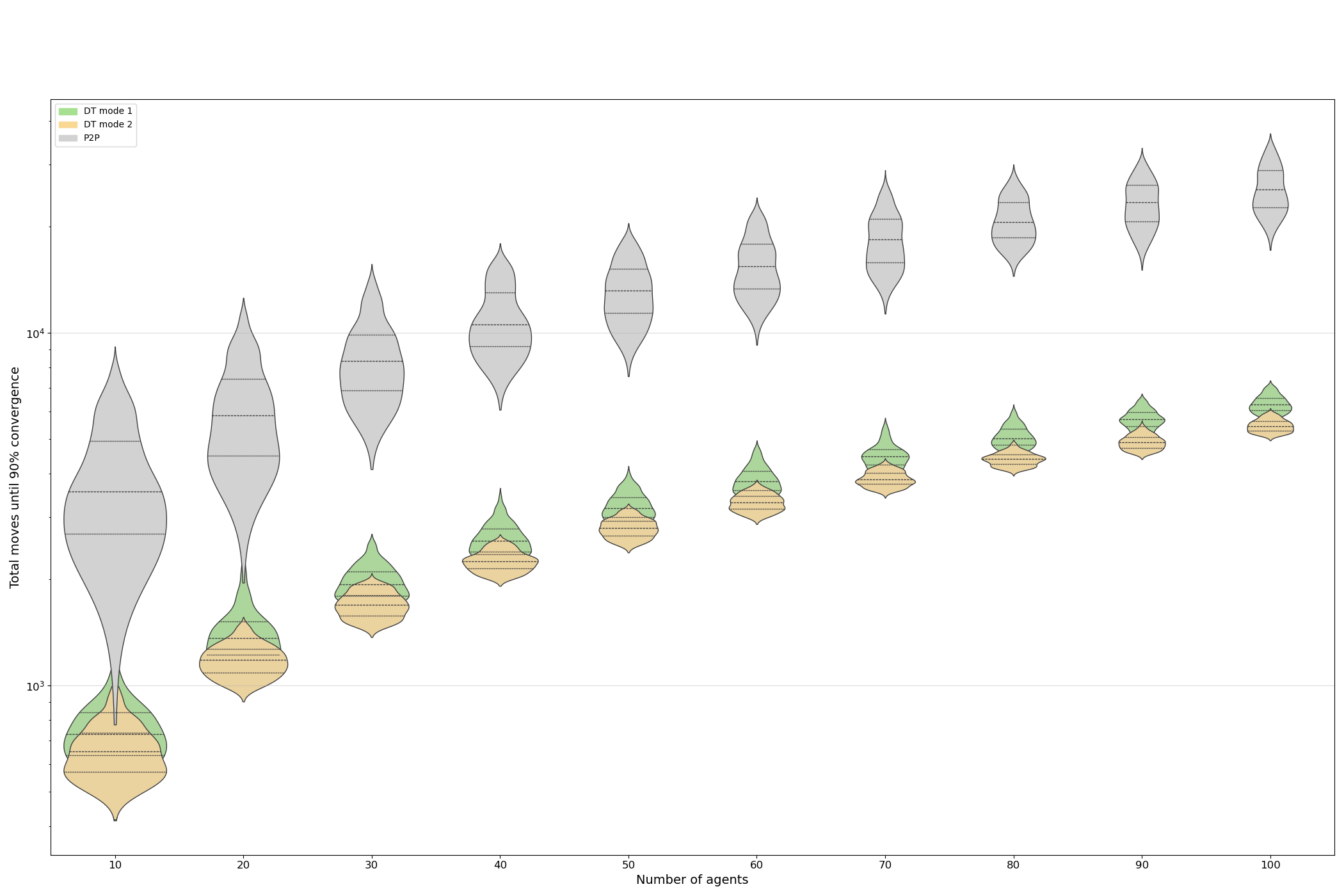}
		\caption{
			Total agent moves until 90\% convergence in three modes:
			(1) DT mode~1 -- DT assisted PSO with a single data exchange per iteration, 
			(2) DT mode~2 -- DT assisted PSO with adaptive upload/download frequency based on the P2P reference, and 
			(3) P2P -- PSO without DT support. 
		}
		\label{fig:LoopCovMoves}
	\end{figure*}

\begin{figure*}[!htpb]
		\centering
		\includegraphics[width=.9\linewidth]{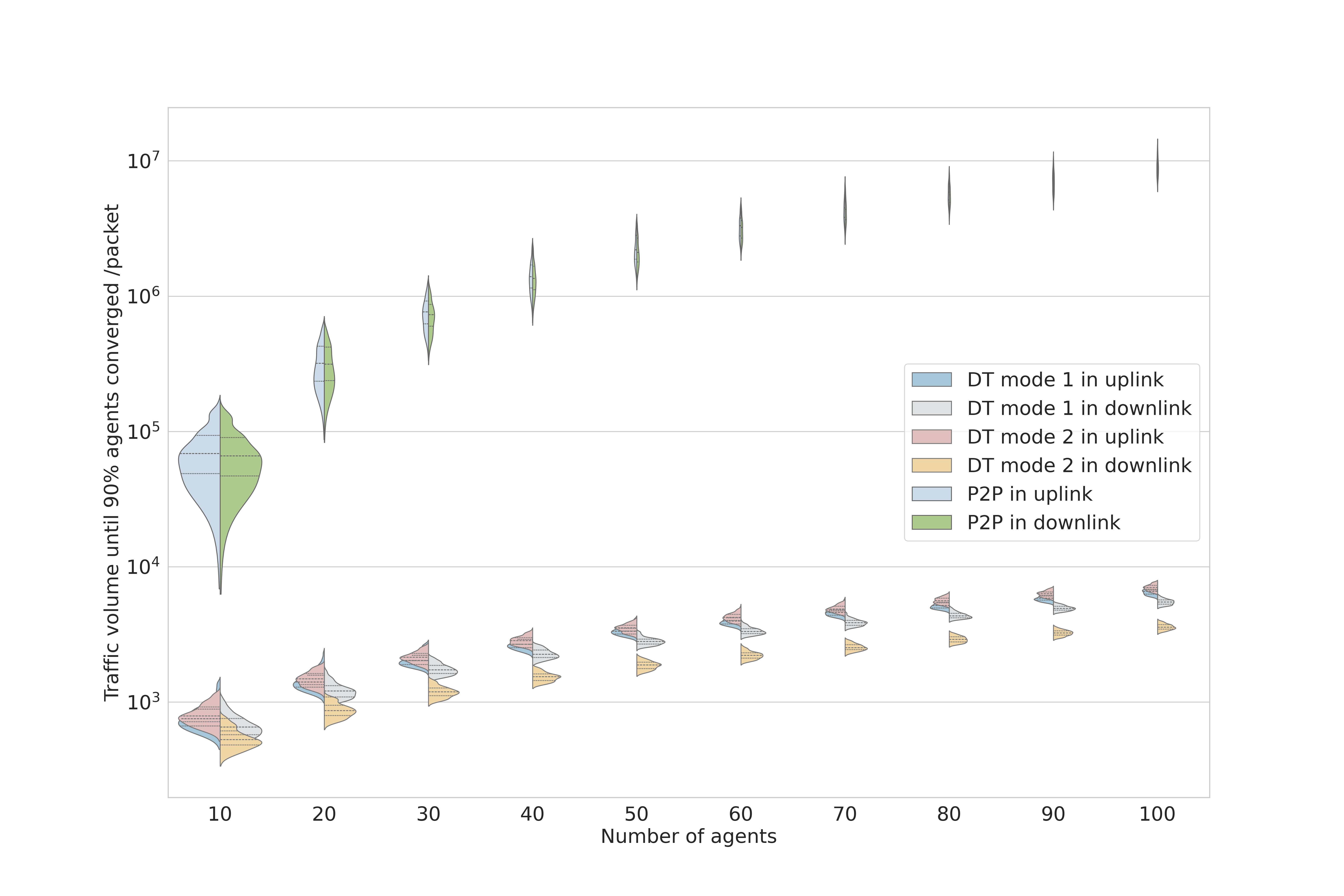}
		\caption{
			Uplink and downlink traffic in three modes:
			(1) DT mode~1 -- DT assisted PSO with a single data exchange attempt per iteration, 
			(2) DT mode~2 -- DT assisted PSO with adaptive upload/download frequency based on the P2P reference, and 
			(3) P2P -- PSO without DT support. 
		}
		\label{fig:SimResult}
	\end{figure*}

\begin{figure*}[!htpb]
		\centering
		\includegraphics[width=.9\linewidth]{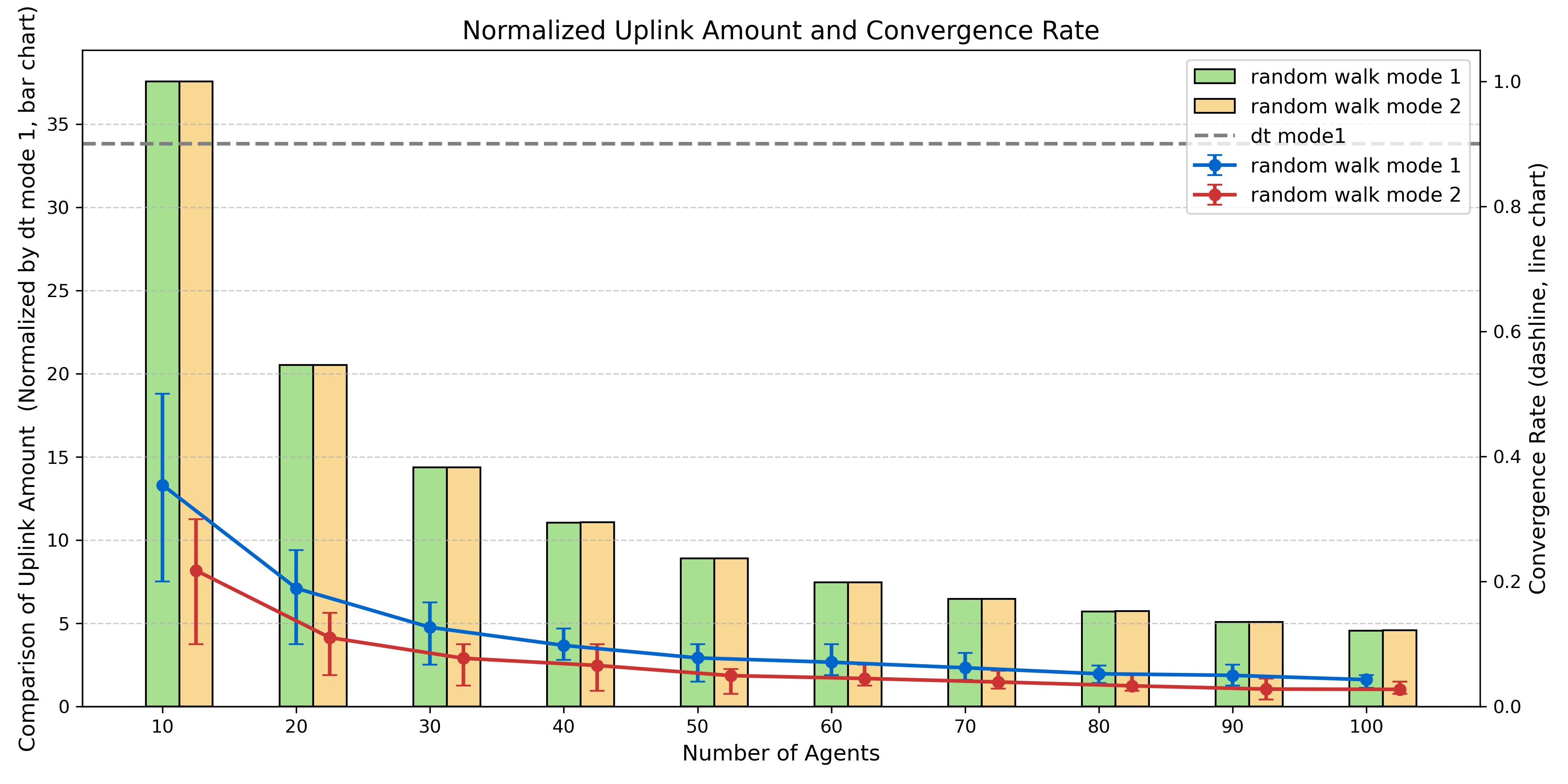}
		\caption{Convergence rate and uplink consumption in three modes:
			(1) Random Walk mode~1, 
			(2) Random Walk mode~2, and 
			(3) DT mode~1 (DT assisted PSO with single data exchange per iteration).
			For each agent group (10--100 agents), the uplink consumption required for DT mode~1 to reach 90\% convergence serves as the reference. 
			The uplink usage of Random Walk modes~1 and~2 is normalized by this reference to highlight their relative communication overhead. 
		}
		\label{fig:CovRate}
	\end{figure*}

	The system is tested with different agent numbers ranging from 10 to 100, with 300 independent Monte Carlo simulation runs conducted for each setup, as shown in Figs.~\ref{fig:LoopCovMoves}--\ref{fig:CovRate}. The results are visualized using violin plots. \revise{The violin plots report the statistical distribution of the results over the Monte Carlo runs rather than a single averaged outcome.
	}Figs.~\ref{fig:LoopCovMoves} and \ref{fig:SimResult} present performance metrics when agent convergence reaches 90\% or higher. \revise{The 90\% convergence threshold is adopted to represent stable swarm localization performance while avoiding excessive sensitivity to a small number of isolated agents affected by poor communication or local obstacle constraints.
	}
	
	As demonstrated in Fig. \ref{fig:LoopCovMoves}, achieving equivalent 90\% convergence requires significantly more movement operations across all agents in P2P mode compared to DT modes. This substantial difference indicates that P2P mode incurs considerably higher temporal and energy costs for \ac{uav}s to achieve comparable performance outcomes to DT approaches. The results underscore the efficiency advantages of the proposed DT framework in terms of both operational time and energy consumption.
	
	Fig.~\ref{fig:SimResult} shows that traffic volumes in both the uplink and downlink increase as the number of agents rises, although at a gradually slowing rate. A clear comparison among the three modes shows that the consumption of communication resources in both \ac{dt} mode 1 and \ac{dt} mode 2 is significantly lower than that in the \ac{p2p} mode. In \ac{dt} mode, the transmissions in the downlink are less than in the uplink, which is closely related to the simulation mechanism. In this mode, if an agent fails to upload, it means the server does not know the current position of the agent and will not send data to the agent. In this case, only the uplink transmission count accumulates. Another reason is that in \ac{dt} mode, any agent reaching the leak source can cease uploading its location information. However, in \ac{p2p} mode, even if an agent reaches the leak source position, it still needs to continuously upload its location information so that other agents that have not reached it can obtain the current \ac{gbest} of the swarm.
	
	It is worth noting in Fig.~\ref{fig:SimResult} that the disparity between transmissions in the uplink and downlink in \ac{dt} mode 2 is notably larger than that in \ac{dt} mode 1. This is due to the data retransmission mechanism in \ac{dt} mode 2. In areas with good signal quality, agents typically achieve successful transmission without exhausting their retry attempts, resulting in differences between upload and download counts that approximate the theoretical expectations for that region. However, in areas with poor signal quality, the difference between upload and download counts increases. In poor signal areas, agents in \ac{dt} mode 2 frequently experience initial upload failures, prompting up to $n-2$ retransmission attempts, while the \ac{mec} server can only establish downlink connections with agents that have successfully completed their uploads. \revise{It is also not difficult to understand why transmissions in the uplink in \ac{dt} mode 2 are higher than those in \ac{dt} mode 1, while transmissions in the downlink are lower. Although \ac{dt} mode 2 increases the number of uplink attempts due to the retransmission mechanism, its cumulative downlink count can become lower than that of \ac{dt} mode 1. This seemingly counter intuitive behavior is caused by the closed loop nature of swarm navigation.
		
		The cumulative downlink count can be interpreted as
		\begin{equation}
			D_{\mathrm{total}}=\sum_{t=1}^{T} D(t),
		\end{equation}
		where $D(t)$ denotes the number of downlink transmissions during coordination round $t$, and $T$ is the total number of coordination rounds before convergence.
		
		In \ac{dt} mode 2, repeated upload attempts increase the probability that UAVs obtain updated global best information in poor connectivity regions. As a result, UAVs can leave unfavorable regions earlier and the swarm reaches convergence with fewer coordination rounds. Therefore, even though the downlink opportunity per coordination round may increase, the cumulative downlink count can decrease because the overall mission duration is shortened.
	}
	
	\begin{figure*}[!htpb]
		\centering
		\includegraphics[width=.75\linewidth]{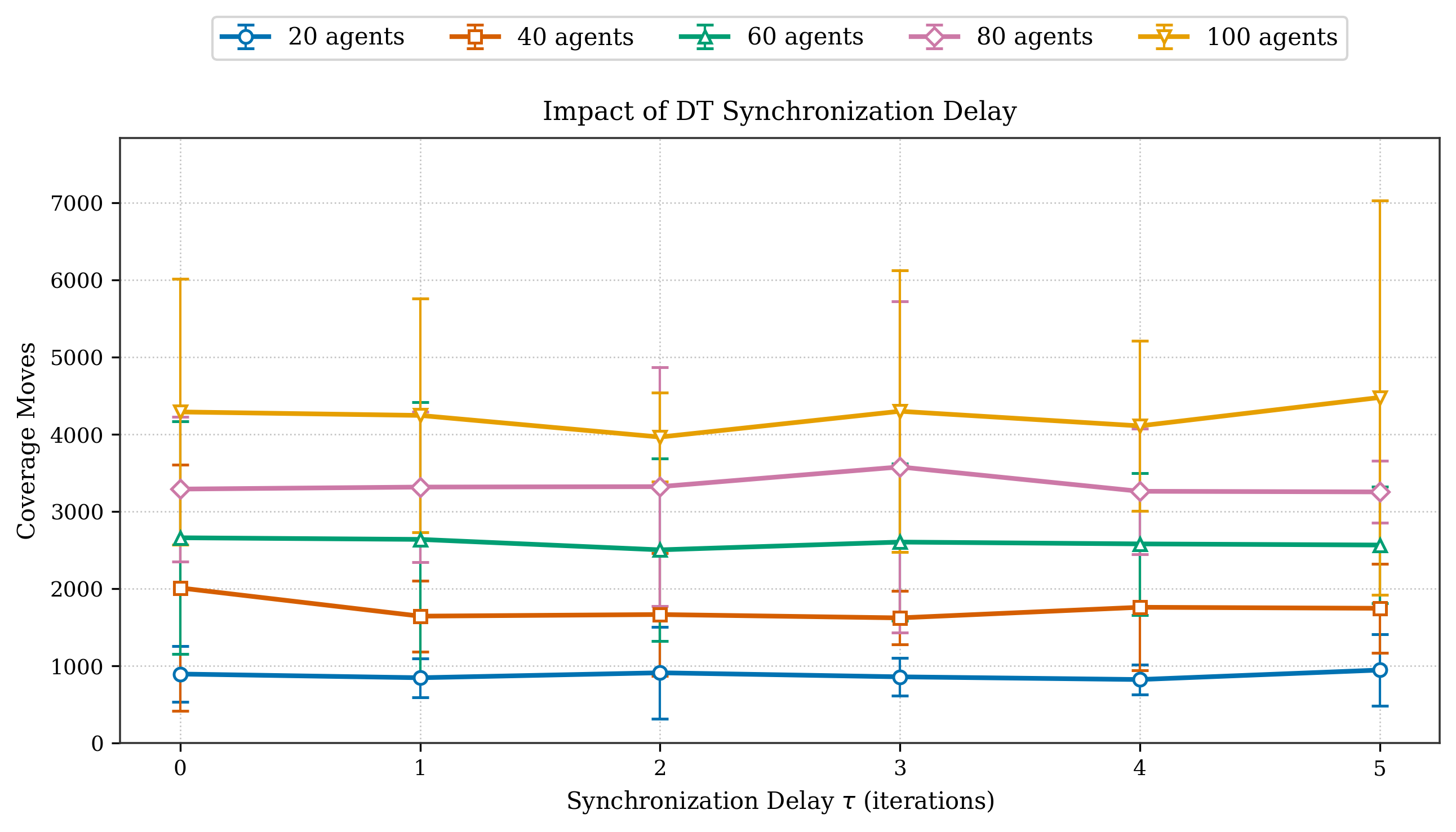}
		\caption{Impact of DT synchronization delay on navigation performance. Average coverage moves required to achieve 90\% convergence are reported for different swarm sizes under bounded synchronization delays ($\tau=0$--$5$).}
		\label{fig:DTdelay}
	\end{figure*}
	
	Fig.~\ref{fig:CovRate} compares random walk modes 1 and 2 with DT mode 1 in terms of uplink transmission frequency and convergence rate. The DT mode 1 data represents uplink consumption required to achieve 90\% convergence for corresponding agent quantities. The bar chart illustrates the multiplicative factor by which random walk modes exceed DT mode 1's uplink usage, while the line chart displays convergence rates. The random walk approach assumes that the \ac{mec} infrastructure contains only agent digital twins, excluding obstacle digital twins, thereby limiting the \ac{mec} server's role to \ac{gbest} selection and distribution. When agents become trapped near obstacles, they must rely solely on inherent PSO algorithms and random walk strategies for escape. The results demonstrate significant performance disparities: for instance, with 40 agents, random walk mode consumes approximately 11 times the uplink resources required by DT mode 1 to achieve 90\% convergence, yet achieves less than 20\% convergence rate. The analysis shows that random-walk PSO approaches without obstacle DT and DT-assisted path planning consume far more communication resources and achieve notably lower convergence rates.
	
	\revise{
		Figs.~\ref{fig:LoopCovMoves} through \ref{fig:CovRate} demonstrate that integrating digital twin technology significantly enhances PSO-based swarm coordination in obstacle rich environments, enabling more efficient leak source localization while conserving communication resources.
	}
	
	\revise{
		\subsection{Impact of DT Synchronization Delay}
		
		To evaluate the robustness of the proposed framework under DT synchronization latency, we conduct an additional sensitivity study by introducing a bounded delay $\tau$ into the MEC-hosted DT update process. The MEC server determines the global best position based on delayed DT states, while UAVs continue local PSO updates and random exploration using the latest available coordination information.
		
		To directly address the DT latency concern while limiting additional computational cost, this experiment focuses on the representative DT mode 1 configuration. The synchronization delay is varied as $\tau \in {0,1,2,3,4,5}$ iterations.
		
		As shown in Fig.~\ref{fig:DTdelay}, the average coverage moves remain relatively stable across all tested delay values and swarm sizes. The limited variation indicates that moderate DT synchronization latency has only a minor impact on the proposed DT-assisted swarm coordination framework. A possible explanation is that the global best position typically evolves over multiple coordination rounds rather than changing abruptly at every iteration. Consequently, bounded synchronization delays introduce only limited deviations between the DT state and the physical swarm state, resulting in negligible performance degradation.
	}

	\section{Discussions}\label{sec:discussions}
	The proposed method enables multiple agents to safely converge to unknown leakage points in obstacle laden environments while saving considerable channel resources compared to the \ac{p2p} mode. \revise{This communication efficiency is primarily achieved through centralized DT assisted synchronization and reduced inter agent communication dependency.} For instance, in the \ac{dt} mode, \revise{once at least one agent successfully reaches the target region, the corresponding target position can be preserved through the DT assisted coordination framework. Even if the agent subsequently leaves the target region, the stored information can still be propagated to other agents whenever communication connectivity becomes available.} Furthermore, this method is device agnostic, allowing agents to be \ac{uav}s, \ac{agv}s, or other \ac{iot} devices.
	
	In the proposed system, agents only need to share their position and $d_i$ with the MEC server rather than broadcasting to all other agents, reducing the exposure of individual location data. The \ac{gbest} position distributed by the MEC is not bound to any specific agent ID, further protecting individual agent privacy. Since the MEC server typically maintains higher security levels and trust measures, privacy protection is enhanced by reducing the attack surface to a single, well secured point rather than exposing location data across multiple peer-to-peer connections. \revise{This DT assisted coordination approach reduces the exposure of sensitive positional information compared with fully distributed peer-to-peer information exchange while maintaining effective swarm coordination.
	}
	
	When the \ac{mec} server receives a help request signal from any agent, the path planning algorithm can fully utilize the known map, which corresponds to the digital twin, for path planning. \revise{Centralizing path planning at the MEC server avoids requiring all agents to continuously execute local obstacle aware path planning algorithms during normal swarm exploration.} Since path planning is only activated when agents become trapped, this approach potentially conserves more resources than deploying path planning capabilities on each individual agent. Additionally, the computational power and processing speed of the \ac{mec} server generally surpass those of individual agents.
	
	\revise{
		Although the current study focuses on static obstacle abstractions, UAV swarms possess additional maneuverability in the vertical dimension compared with ground robotic systems. \revise{The current framework primarily focuses on communication aware swarm coordination rather than low level UAV flight control or aerodynamic modeling.
		} Therefore, certain obstacle interactions can be alleviated through altitude adjustment strategies, reducing the probability of complete path blockage scenarios. 
	}\revise{Future work will further investigate more complex dynamic environments, including moving obstacles and inter agent collision avoidance.
	} These enhancements would provide a more comprehensive understanding of practical applications. \revise{Additional DT synchronization delay experiments indicate that moderate synchronization latency does not critically degrade swarm convergence performance, suggesting a certain degree of robustness of the proposed framework against delayed DT updates.} Future research should also explore optimizing signal propagation in complex environments. This includes forming realistic radio source distributions based on obstacle layouts to better reflect practical conditions. Additionally, optimizing the placement of wireless access points, predicting signal strength and quality, and modeling how wireless signals propagate in complex environments will enhance our understanding of practical applications. \revise{
		These extensions are particularly relevant for large scale industrial inspection, hazardous gas leakage localization, and emergency response scenarios operating under communication constrained environments.
	}
	\section{Conclusion}\label{sec:conclusion}
	\revise{This study demonstrates that DT-assisted swarm coordination can effectively support multi-agent leak source localization in communication constrained and obstacle laden environments.} The proposed method conserves both communication and energy resources by reducing the number of necessary transmissions. \revise{Local UAV decision making based on previously received global coordination information reduces unnecessary raw data transmission and communication overhead.} \revise{The proposed framework provides a scalable foundation for future communication aware swarm coordination systems involving UAVs, AGVs, and heterogeneous IoT assisted autonomous platforms.} \revise{Additional synchronization delay analysis further indicates that the proposed framework maintains stable swarm coordination performance under bounded DT synchronization latency.
	}

	\ifCLASSOPTIONcaptionsoff
	\newpage
	\fi

	\bibliographystyle{IEEEtran}
	\bibliography{CiteTheses}

\begin{thebibliography}{10}
\providecommand{\url}[1]{#1}
\csname url@samestyle\endcsname
\providecommand{\newblock}{\relax}
\providecommand{\bibinfo}[2]{#2}
\providecommand{\BIBentrySTDinterwordspacing}{\spaceskip=0pt\relax}
\providecommand{\BIBentryALTinterwordstretchfactor}{4}
\providecommand{\BIBentryALTinterwordspacing}{\spaceskip=\fontdimen2\font plus
\BIBentryALTinterwordstretchfactor\fontdimen3\font minus \fontdimen4\font\relax}
\providecommand{\BIBforeignlanguage}[2]{{%
\expandafter\ifx\csname l@#1\endcsname\relax
\typeout{** WARNING: IEEEtran.bst: No hyphenation pattern has been}%
\typeout{** loaded for the language `#1'. Using the pattern for}%
\typeout{** the default language instead.}%
\else
\language=\csname l@#1\endcsname
\fi
#2}}
\providecommand{\BIBdecl}{\relax}
\BIBdecl

\bibitem{9349624}
W.~Jiang, B.~Han, M.~A. Habibi, and H.~D. Schotten, ``The road towards {6G}: {A} comprehensive survey,'' \emph{IEEE Open Journal of the Communications Society}, vol.~2, pp. 334--366, 2021.

\bibitem{chowdhury20206g}
M.~Z. Chowdhury, M.~Shahjalal, S.~Ahmed, and Y.~M. Jang, ``{6G} wireless communication systems: Applications, requirements, technologies, challenges, and research directions,'' \emph{IEEE Open Journal of the Communications Society}, vol.~1, pp. 957--975, 2020.

\bibitem{khan2022digital}
L.~U. Khan, W.~Saad, D.~Niyato, Z.~Han, and C.~S. Hong, ``Digital-twin-enabled {6G}: Vision, architectural trends, and future directions,'' \emph{IEEE Communications Magazine}, vol.~60, no.~1, pp. 74--80, 2022.

\bibitem{han2023digital}
B.~Han, M.~A. Habibi, B.~Richerzhagen, K.~Schindhelm, F.~Zeiger, F.~Lamberti, F.~G. Prattic{\`o}, K.~Upadhya, C.~Korovesis, I.-P. Belikaidis \emph{et~al.}, ``Digital twins for industry 4.0 in the {6G} era,'' \emph{IEEE Open Journal of Vehicular Technology}, 2023.

\bibitem{9923927}
N.~P. Kuruvatti, M.~A. Habibi, S.~Partani, B.~Han, A.~Fellan, and H.~D. Schotten, ``Empowering {6G} communication systems with digital twin technology: A comprehensive survey,'' \emph{IEEE Access}, vol.~10, pp. 112\,158--112\,186, 2022.

\bibitem{DenKr_DigitalTwin_MathModel}
D.~Krummacker, M.~Reichardt, C.~Fischer, and H.~D. Schotten, ``\BIBforeignlanguage{English}{\begingroup {Digital twin development: Mathematical modeling}\endgroup},'' in \emph{\BIBforeignlanguage{English}{ICPS 2023 -- 6th IEEE International Conference on Industrial Cyber-Physical Systems}}.\hskip 1em plus 0.5em minus 0.4em\relax IEEE, 5 2023, p.~8.

\bibitem{bonabeau1999swarm}
E.~Bonabeau, M.~Dorigo, and G.~Theraulaz, \emph{Swarm intelligence: {From} natural to artificial systems}.\hskip 1em plus 0.5em minus 0.4em\relax Oxford university press, 1999.

\bibitem{martens2011editorial}
D.~Martens, B.~Baesens, and T.~Fawcett, ``Editorial survey: {Swarm} intelligence for data mining,'' \emph{Machine Learning}, vol.~82, pp. 1--42, 2011.

\bibitem{YHKS2022massive}
S.~Yuan, B.~Han, D.~Krummacker, and H.~D. Schotten, ``Massive twinning to enhance emergent intelligence,'' in \emph{2nd International Workshop on Distributed and Intelligent Systems (DistInSys 2022)}, 2022, pp. 1--4.

\bibitem{mcguire2019minimal}
K.~N. McGuire, C.~De~Wagter, K.~Tuyls, H.~J. Kappen, and G.~C. de~Croon, ``Minimal navigation solution for a swarm of tiny flying robots to explore an unknown environment,'' \emph{Science Robotics}, vol.~4, no.~35, p. eaaw9710, 2019.

\bibitem{ducatelle2014cooperative}
F.~Ducatelle, G.~A. Di~Caro, A.~F{\"o}rster, M.~Bonani, M.~Dorigo, S.~Magnenat, F.~Mondada, R.~O’Grady, C.~Pinciroli, P.~R{\'e}tornaz \emph{et~al.}, ``Cooperative navigation in robotic swarms,'' \emph{Swarm Intelligence}, vol.~8, no.~1, pp. 1--33, 2014.

\bibitem{kennedy1995particle}
J.~Kennedy and R.~Eberhart, ``Particle swarm optimization,'' in \emph{Proceedings of ICNN'95 - International Conference on Neural Networks}, vol.~4, 1995, pp. 1942--1948.

\bibitem{DP1985generalized}
\BIBentryALTinterwordspacing
R.~Dechter and J.~Pearl, ``Generalized best-first search strategies and the optimality of {A*},'' \emph{Association for Computing Machinery}, vol.~32, no.~3, p. 505–536, jul 1985. [Online]. Available: \url{https://doi.org/10.1145/3828.3830}
\BIBentrySTDinterwordspacing

\bibitem{Dijkstra1959note}
E.~W. Dijkstra, ``A note on two problems in connexion with graphs.'' \emph{Numerische Mathematik}, vol.~1, pp. 269--271, 1959.

\end{thebibliography}
	
	\newpage
	
	\begin{IEEEbiography}[{\includegraphics[width=1in,height=1.25in,clip,keepaspectratio]{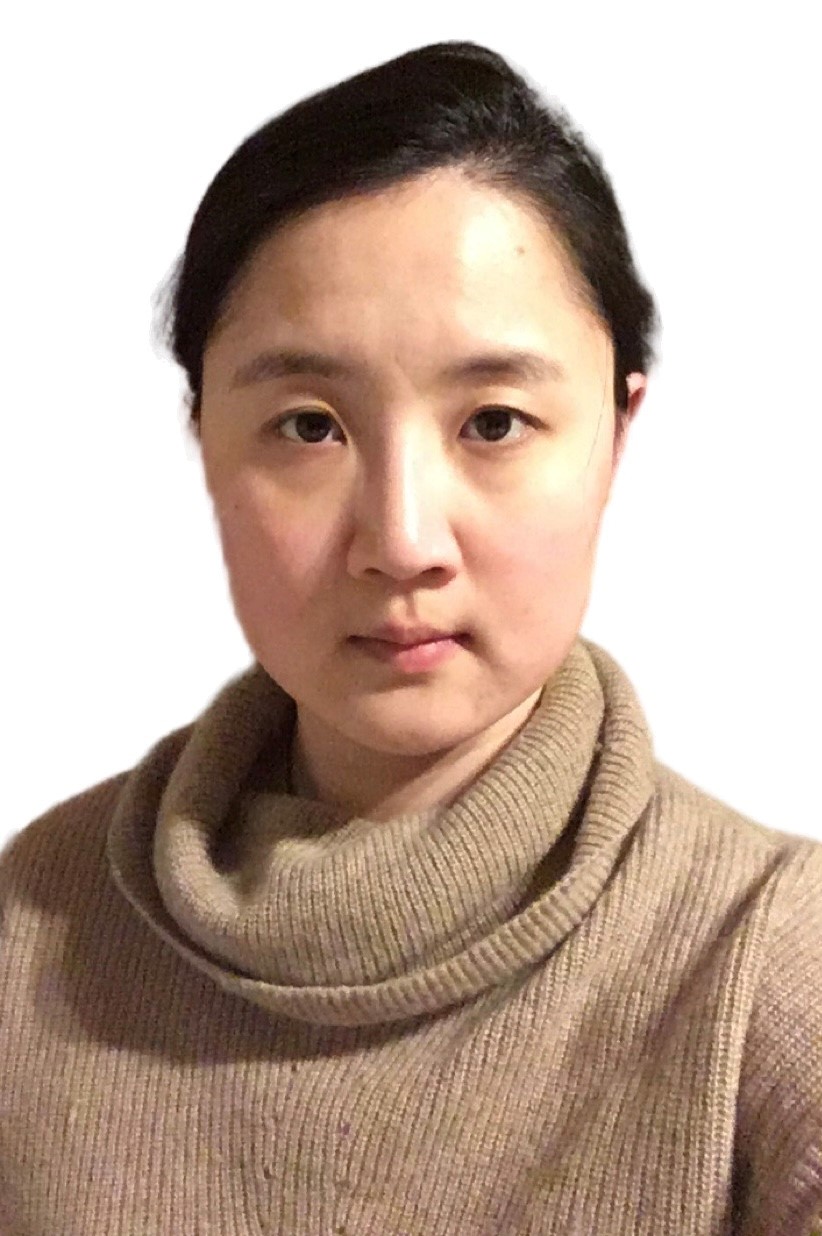}}]{Siyu Yuan} received her M.Sc. degree in Automation and Robotics from the University of Dortmund in 2019, Germany. Since then, she has been a researcher with the Division of Wireless Communications and Radio Positioning (WiCON) at RPTU Kaiserslautern-Landau. Her main research interests are in drone communication, swarm intelligence with digital twinning, distributed simultaneous localization and mapping, machine learning in robotics.
	\end{IEEEbiography}
	\vspace{-1.0\baselineskip}
	
	\begin{IEEEbiography}[{\includegraphics[width=1in,height=1.25in,clip,keepaspectratio]{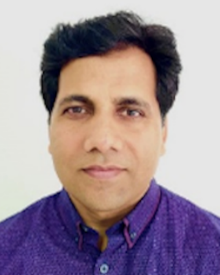}}]{Khurshid Alam} received his M. Sc. degree in Applied Computer Science from the University of Kaiserslautern, Germany in 2018. Since then, he has been a researcher with the Intelligent Networks research group at the German Research Center for Artificial Intelligence (DFKI) GmbH, Germany. His main research interests include industrial wireless communication, software defined networking, radio access network architecture, radio resource management, artificial intelligence, network function virtualization, and network slicing.
	\end{IEEEbiography}
	\vspace{-1.0\baselineskip}
	
	\begin{IEEEbiography}[{\includegraphics[width=1in,height=1.25in,clip,keepaspectratio]{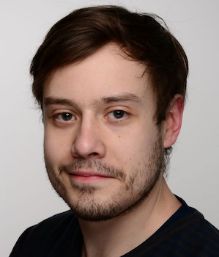}}]{Dennis Krummacker} received the M.Sc. degree in Electrical and Computer Engineering from the University of Kaiserslautern in 2016. Since 2016, he has been a Researcher at the German Research Center for Artiﬁcial Intelligence GmbH (DFKI) in the Department of Intelligent Networks, where he is coordinating the research areas regarding Industrial Communication and Intelligent Infrastructures. His main research interests include Network Management and Orchestration, Softwarization, Virtualization, Software-defined Networking (SDN), and Network Function Virtualization (NFV).
	\end{IEEEbiography}
	\vspace{-1.0\baselineskip}
	
	\begin{IEEEbiography}[{\includegraphics[width=1in,height=1.25in,clip,keepaspectratio]{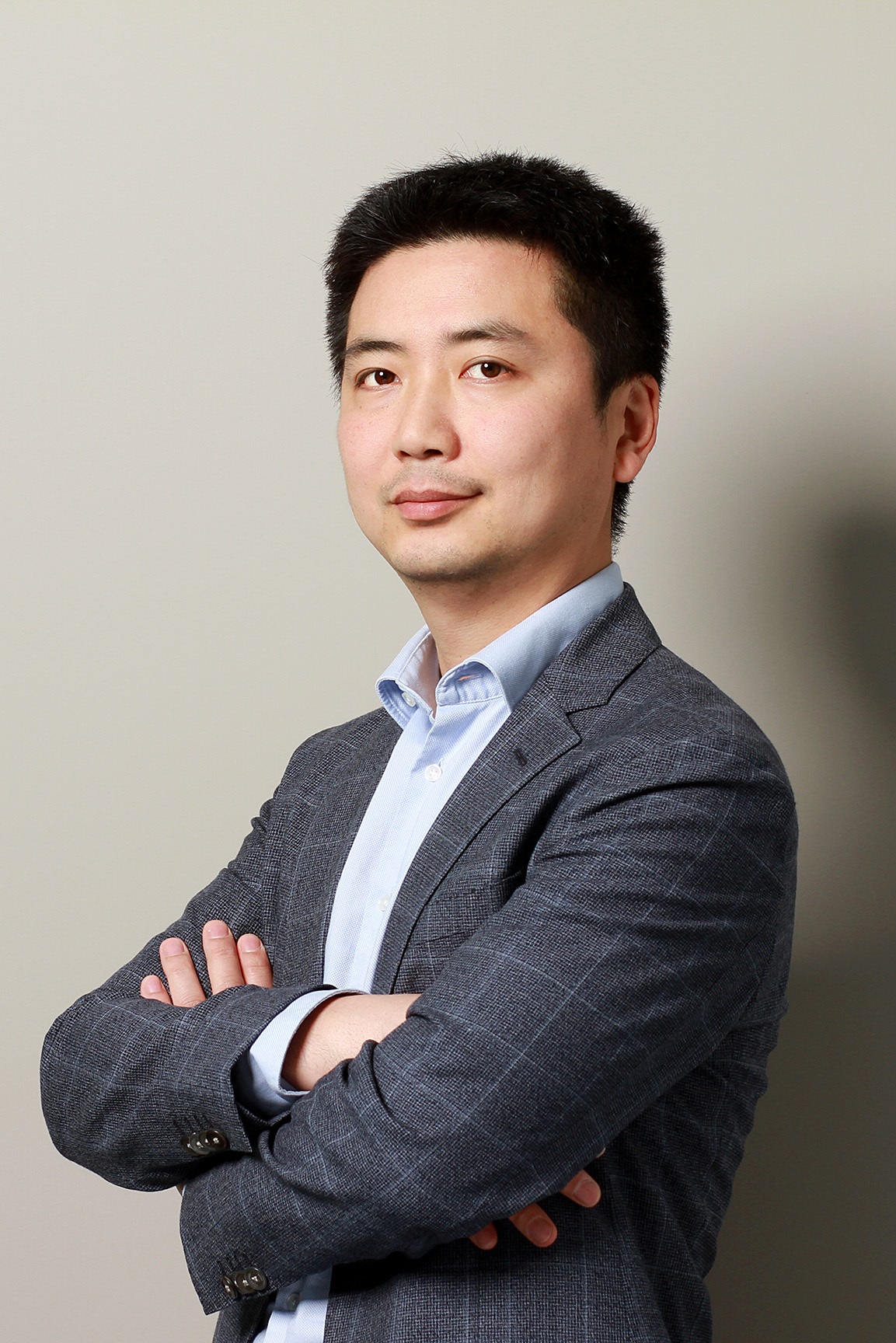}}]{Bin Han} received the B.E. degree from Shanghai Jiao Tong University, China, in 2009, the M.Sc. degree from the Darmstadt University of Technology, Germany, in 2012, and the Ph.D. (Dr.-Ing.) degree from the Karlsruhe Institute of Technology, Germany, in 2016. He joined RPTU in July 2016, where he has been working as Postdoctoral Researcher and Senior Lecturer. In November 2023, he was granted the teaching license (Venia Legendi) by RPTU, and there with became a Privatdozent. He is the author of two books, six book chapters, and over 90 research papers. He has participated in multiple EU FP7, Horizon 2020, and Horizon Europe research projects. He is Editor of IEEE Wireless Communications Letters and Editorial Board Member of Network. He has served in organizing committee and/or TPC for IEEE GLOBECOM, IEEE ICC, EuCNC, European Wireless, and ITC. He is actively involved in the IEEE Standards Association Working Groups P1955, P2303, P3106, and P3454.
	\end{IEEEbiography}
	\vspace{-1.0\baselineskip}
	
	\begin{IEEEbiography}[{\includegraphics[width=1in,height=1.25in,clip,keepaspectratio]{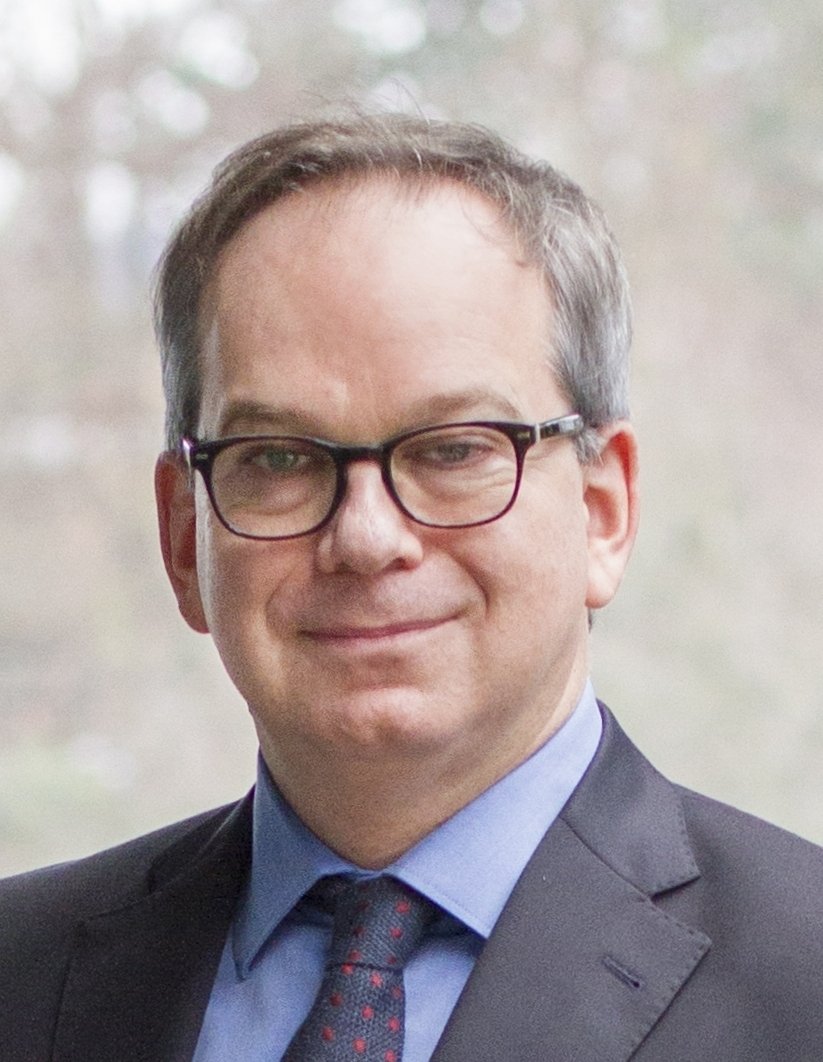}}]{Hans D. Schotten} received the Ph.D. degree from the RWTH Aachen University of Technology, Germany, in 1997. From 1999 to 2003, he worked for Ericsson. From 2003 to 2007, he worked for Qualcomm. He became manager of a R\&D group, Research Coordinator for Qualcomm Europe, and Director for Technical Standards. In 2007, he accepted the offer to become a full professor at the University of Kaiserslautern. In 2012, he - in addition - became the scientific director of the German Research Center for Artificial Intelligence (DFKI) and head of the Department for Intelligent Networks. Professor Schotten served as dean of the Department of Electrical Engineering of the University of Kaiserslautern from 2013 until 2017. Since 2018, he is chairman of the German Society for Information Technology and a member of the Supervisory Board of the VDE. He is the author of more than 200 papers and participated in 30+ European and national collaborative research projects.
	\end{IEEEbiography}
	
	\EOD
	
\end{document}